\documentclass[pdflatex,sn-mathphys-num,referee]{sn-jnl}% Math and Physical Sciences Numbered Reference Style

\usepackage{graphicx}%
\usepackage{multirow}%
\usepackage{multicol}
\usepackage{amsmath,amssymb,amsfonts}%
\usepackage{amsthm}%
\usepackage{mathrsfs}%
\usepackage[title]{appendix}%
\usepackage[table,xcdraw]{xcolor}%
\usepackage{textcomp}%
\usepackage{manyfoot}%
\usepackage{booktabs}%
\usepackage{algorithm}%
\usepackage{algorithmicx}%
\usepackage{algpseudocode}%
\usepackage{listings}%
\usepackage{subcaption}
\usepackage{adjustbox}
\usepackage{caption}
\usepackage{tabu}
\usepackage{longtable}
\usepackage{footnote}
\usepackage{hyperref}
\usepackage{comment}
\usepackage{threeparttable}
\usepackage{makecell}
\usepackage{mathtools}
\usepackage{float}
\usepackage{array}
\usepackage{tabularx}
\usepackage{lineno}
\theoremstyle{thmstyleone}%

\theoremstyle{thmstyletwo}%

\theoremstyle{thmstylethree}%
\usepackage{setspace}

\begin{document}

\title[Improving MSA Estimation through Adaptive Weight Vectors in MOEA/D]{Improving MSA Estimation through Adaptive Weight Vectors in MOEA/D}
%%=============================================================%%
%% Author Information
%%=============================================================%%

\author[1]{\fnm{Saem} \sur{Hasan}}\email{saemhasan@cse.buet.ac.bd}

\author[2]{\fnm{Muhammad Ali} \sur{Nayeem}}\email{nayeem@qu.edu.sa}

\author*[1]{\fnm{M. Sohel} \sur{Rahman}}\email{sohel.kcl@gmail.com}

\affil[1]{\orgdiv{Department of Computer Science and Engineering}, \orgname{Bangladesh University of Engineering and Technology}, \orgaddress{\city{Dhaka}, \postcode{1205}, \country{Bangladesh}}}

\affil[2]{\orgdiv{Department of Computer Engineering}, \orgname{Qassim University}, \orgaddress{\city{Buraydah}, \postcode{52571}, \country{Saudi Arabia}}}

%%==================================%%
%% Abstract
%%==================================%%

\abstract{Accurate phylogenetic inference from biological sequences depends critically on the quality of multiple sequence alignments, yet optimal alignment for many sequences is computationally intractable and sensitive to scoring choices. In this work we introduce \textbf{MOEA/D-ADF}, a novel variant of MOEA/D that adaptively adjusts subproblem weight vectors based on fitness variance to improve the exploration--exploitation trade-off. We combine MOEA/D-ADF with PMAO (PASTA with many application-aware optimization criteria) to form \textbf{PMAO\texttt{++}}: PMAO-generated solutions are used to seed MOEA/D-ADF, which then evolves a population using 30 weight vectors to produce a diverse ensemble of alignment--tree pairs. PMAO\texttt{++} outperforms the original PMAO on a majority of benchmark cases, achieving better false-negative (FN) rates on 12 of 17 BAliBASE-derived datasets and producing superior best-case trees, including several instances with zero FN rate. Beyond improving single best alignments, the rich set of alignment--tree pairs produced by PMAO\texttt{++} is especially valuable for downstream summary methods (for example, consensus and summary-tree approaches), allowing more robust phylogenetic inference by integrating signal across multiple plausible alignments and trees. Certain dataset features (such as large terminal N/C extensions found in the RV40 group) remain challenging, but overall PMAO\texttt{++} demonstrates clear advantages for sequence-based phylogenetic analysis. Future work will explore parameter tuning, larger benchmark suites, and tighter integration with summary-tree pipelines to further enhance applicability for biological sequence studies.}

% \keywords{Multi-objective optimization, MOEA/D, Multiple sequence alignment, Phylogenetic inference.}

\maketitle

\section{Introduction}\label{sec:introduction}

Optimization problems are found everywhere in real-life scenarios. When there are several objective functions that require simultaneous optimization, and improving the performance of one objective typically has an adverse effect on one or more other objectives, the problem is referred to as a multi-objective optimization problem (MOP). MOPs with more than three objectives are known as many-objective optimization problems (MaOPs) \cite{li2015many}.

Evolutionary algorithms (EAs) are widely used to solve MOPs and MaOPs. Multi-Objective Evolutionary Algorithms based on Decomposition (MOEA/D)\cite{moead} represent a powerful class of optimization techniques designed to address complex multi-objective optimization problems. Although traditional implementations of MOEA/D are effective, they can face difficulties in exploring the search space. As a result, suboptimal solutions may be generated, especially in cases where the optimization landscape is complex and dynamic. Overcoming these challenges requires the development of novel strategies that can improve the trade-off between exploration and exploitation, and enhance the convergence behavior of MOEA/D.

Recent research has focused on enhancing MOEA/D (Multi-Objective Evolutionary Algorithm Based on Decomposition) with adaptive mechanisms that dynamically adjust the search strategy. One method proposed in \cite{qi2014moea} involves periodically adjusting the weights of the subproblems. This adaptive redistribution of subproblem weights results in better uniformity of solutions. Another strategy proposed in \cite{dong2020moea} is a self-adaptive weight vector adjustment method based on chain segmentation. It aims to improve the performance of MOEA/D. Additionally, a new strategy called MOEA/D-ANA (MOEA/D with adaptive neighborhood size adjustment) was presented in \cite{xu2021adaptive} to increase diversity. MOEA/D-AWS \cite{gu2024moea} uses subproblem evolutionary matrix similarity to adjust weight vectors. Lastly, MaOEA/D-AEW \cite{sun2024maoea} proposes a MOEA/D with adaptive external population-guided weight vector adjustment.

% \cite{he2023moea}population agents stored in the external set are reused for updating the weight adaptively during the evolutionary process. mainly updated the weight vectors adaptively. 

Sequences play a crucial role in biology. DNA stores and transmits information, RNA is transcribed from DNA, and proteins are translated from RNA sequences. Proteins are the building blocks of living organisms. Multiple Sequence Alignment (MSA) is a tool used to compare sequences and is used in bioinformatics pipelines for estimating phylogenies, predicting the structure/function of peptides, and comparing genomic sequences.

Pairwise alignment problems for two sequences can be solved in quadratic time. However, when we handle many sequences, finding the optimal alignment becomes NP-hard, even with reasonable scoring schemes \cite{Bonizzoni2001May, Caucchiolo2023Feb}. To solve the Multiple Sequence Alignment problem efficiently, many heuristic and metaheuristic approaches have been implemented, including Clustal Omega \cite{Sievers2011Oct}, T-Coffee and TCS \cite{chang2014tcs}, MAFFT \cite{katoh2002mafft}, MUSCLE \cite{edgar2004muscle}, and others. PASTA \cite{mirarab2015pasta} introduces a new iterative approach to generate an alignment based on a guide tree, which makes it highly scalable and accurate. In each iteration, PASTA improves both the tree and its corresponding alignment.

There is a sufficient amount of literature that shows the multiobjective optimization approach can be highly beneficial in inferring MSAs \cite{ortuno2013optimizing, soto2014multi, abbasi2015local, rubio2016hybrid}, despite the fact that traditionally, MSA methods have been benchmarked on SP score and TC score \cite{warnow2017computational}. We based our experiment on \textit{PASTA with many application-aware optimization criteria for alignment-based phylogeny inference} (\textbf{PMAO}) \cite{pmao}. PMAO solves multiobjective optimization problems by breaking them down into single-objective optimization problems using weight vectors. PMAO uses 30 weight vectors and outputs 30 alignment-tree pairs, each optimal for a specific weight vector.

We have proposed a modified version of MOEA/D, which is MOEA/D with Adaptive Weight vectors based on the variance of Fitness (\textbf{MOEA/D-ADF}). We treated PMAO solutions as the initial population and ran our method MOEA/D-ADF. The main contribution of this project is the proposal of MOEA/D-ADF and combining it with PMAO.  

\section{Background}\label{sec:background}

% \addc{We first briefly review MOEA/D and PMAO in this section. Then we explain our workflow to infer MSAs and then cluster them.}

\subsection{MOEA/D}

In a multi-objective setting, we have several different, often negatively correlated, objectives we want to optimize. Instead of providing one solution, multiobjective optimization algorithms provide a set of solutions, often known as the \emph{pareto front}, which are \emph{dominated} by no other solution. We say a solution $X$ \emph{dominates} another solution $Y$, if $X$ is at least as good as $Y$ in all objectives and strictly better in at least one. Many algorithms for multiobjective optimizations have been proposed (for a detailed discussion, check \cite{Luke2012Dec}). 

A Multiobjective Evolutionary Algorithm Based on Decomposition (MO-EA/D) and its modifications been utilized successfully across many applications, including engineering optimization, network clustering and community detection, neural architecture search, vehicle routing problems etc \cite{moead, moead_survey}. MOEA/D algorithm has two crucial functional building blocks:

\begin{itemize}
    \item \textbf{Decomposition:} We divide the multiobjective optimization problem into several scalar optimization problems. We have used the Weighted Chebyshev approach, where we formulate each scalar optimization problem as:

    $$
    \min_{x \in \Omega} g^{tch} (\mathbf{x}|\mathbf{w}, \mathbf{z}^{*}) = \max_{1 \leq i \leq m} \{w_i|f_i(\mathbf{x}) - {z}_i^*| \} 
    $$

    where $\{f_i(\mathbf{x})\}_{i=1}^N$ are the individual objective functions we seek to optimize, $\Omega$ is the permitted domain, $w_i$ denotes the weight of the $i^{th}$ objective and $z_i^{*} = \min_{x \in \Omega} f_i(x)$ denotes the ideal reference point. 
    \item \textbf{Collaboration: } MOEA/D, a multi-objective optimization approach, differs from traditional methods by concurrently addressing all subproblems using a population-based metaheuristic. It assumes that neighboring subproblems, determined by the Euclidean distance between their weight vectors, share similar characteristics. This collaboration allows solutions within the same neighborhood to exchange elite information, with mating parents selected from the neighborhood for offspring reproduction, potentially leading to the replacement of a subproblem's associated solution if it improves fitness.
\end{itemize}

% \begin{algorithm}
% \caption{Multi-Objective Optimization Algorithm }
% \SetKwInOut{Input}{Input}
% \SetKwInOut{Output}{Output}

% \Input{Initialize a population of solutions $P = \{x^i\}_{i=1}^N$, a set of weight vectors $W = \{w^i\}_{i=1}^N$, and their neighborhood structure. Randomly assign each solution to a weight vector.}
% \Output{The final population of solutions.}
% \While{stopping criteria not met}{
%     \For{$i = 1$ to $N$}{
%         \textbf{Step 2.1:} Randomly select a required number of mating parents from $w^i$'s neighborhood, denoted as $\Theta^{w^i}$.
        
%         \textbf{Step 2.2:} Use crossover and mutation to reproduce offspring $x^c$.
        
%         \textbf{Step 2.3:} Update the subproblems within $\Theta^{w^i}$ by $x^c$ if $g(x^c | w^k, z^*) < g(x^k | w^k, z^*)$, where $w^k \in W$ and $x^k$ is the current solution associated with $w^k$.
%     }
% }
% \caption{Multi-Objective Optimization Algorithm \cite{moead_survey} \label{moead1}}
% \end{algorithm}

\subsection{PMAO}
% sayem
% \addc{@someone, write about PMAO} \\
\textbf{PMAO} (\textbf{P}ASTA with \textbf{M}any \textbf{A}pplication-aware \textbf{O}ptimization criteria for alignment-based phylogeny inference)\cite{nayeem2022pasta} is a method that extends the capabilities of the well-known \textbf{PASTA}\cite{mirarab2015pasta} multiple sequence alignment (MSA) tool. PMAO presents a multi-objective framework by incorporating four application-aware objectives and PASTA's typical maximum likelihood (ML) score. This unique method generates a set of high-quality MSA and tree solutions, overcoming the constraints of utilizing ML score alone. The experimental analysis of PMAO shows that it can generate much better phylogenetic trees than independent PASTA. 

Alongside the ML score, in the PMAO framework, they incorporated the following four simple objective functions that were identified by Nayeem et al.\cite{nayeem2020multiobjective} based on their better correlation to the tree accuracy.

\begin{enumerate}
    \item \textbf{Maximize similarity for columns containing gaps (SIMG):} For each column of the MSA having at least one gap, it calculates the ratio of the most frequent characters. Then all those ratios are added to get the SIMG score.

    \item \textbf{Maximize similarity for columns containing no gaps (SIMNG):} This is similar to SIMG except that it considers those columns of the MSA that do not have any gaps.

    \item \textbf{Maximize sum-of-pairs (SOP):} For each pair of aligned sequences in the MSA, it takes the sum of substitution score for the two aligned characters across all columns using a substitution matrix. The addition of all pairwise scores gives the SOP score. In the PMAO paper, they used the BLOSUM62 substitution matrix\cite{henikoff1992amino}.

    \item \textbf{Minimize the number of gaps (GAP):} This is the summation of the number of gap characters in each aligned sequence. For the sake of uniformity, they converted this score into a maximization criterion.
\end{enumerate}

\section{Methods}\label{sec:methods}

\subsection{MOEA/D-ADF}
% MOEAD-ADF
We have proposed a modified version of MOEA/D, which is \textbf{MOEA/D} with \textbf{A}daptive \textbf{W}eight vectors based on the variance of \textbf{F}itness (\textbf{MOEA/D-ADF}) in the phenotype space. Initially, all the weight vectors have the same weight. After running one iteration for each weight vector, we evaluate the fitness variance of the solutions associated with the weight vector. The weight vector with a higher variance gets a higher weight in the next iteration. The weight vector that produces solutions with more diversity gets to explore more in the search space in the next iteration. We are promoting more exploration of the weight vector by assigning weight based on the solutions' fitness variances. 

% \begin{algorithm}
% \caption{Adaptive Weight Vector Assignment in MOEA/D Algorithm}
% \SetKwInOut{Input}{Input}
% \SetKwInOut{Output}{Output}

% \Input{Initialize a population of solutions $P = \{x^i\}_{i=1}^N$, a set of weight vectors $W = \{w^i\}_{i=1}^N$, and their neighborhood structure. Randomly assign each solution to a weight vector.}
% \Output{The final population of solutions.}
% \While{stopping criteria not met}{
%     \textbf{Step 1:} Assign an equal weight to each weight vector and create a set of weighted weight vectors, WV.\\
    
%     \For{Each Weight Vector, $w_i$ : WV}{
%         \For{$i = 1$ to $N$}{
%             \textbf{Step 2.1:} Randomly select a required number of mating parents from $w^i$'s neighborhood, denoted as $\Theta^{w^i}$.
            
%             \textbf{Step 2.2:} Use crossover and mutation to reproduce offspring $x^c$.
            
%             \textbf{Step 2.3:} Update the subproblems.
%         } 
%     }
%     \textbf{Step 3:} Update WV based on the solutions' fitness variances.
% }
% % \caption{Multi-Objective Optimization Algorithm \cite{moead_survey} 
% % \label{moead}
% \end{algorithm}

\begin{algorithm}[H]
\caption{Adaptive Weight Vector Assignment in MOEA/D Algorithm}
\label{alg:moead-adf}
\begin{algorithmic}[1]
\Require Population of solutions $P = \{x^i\}_{i=1}^N$, set of weight vectors $W = \{w^i\}_{i=1}^N$, neighborhood structure.
\Ensure The final population of solutions..
\State Randomly assign each solution $x^i$ to a weight vector $w^i$.
\While{stopping criteria not met}
    \State \textbf{Step 1:} Assign an equal weight to each weight vector and create the set of weighted weight vectors $WV$.
    \For{each weight vector $w_i \in WV$}
        \For{$i = 1$ to $N$}
            \State \textbf{Step 2.1:} Randomly select the required number of mating parents from the neighborhood of $w^i$, denoted $\Theta^{w^i}$.
            \State \textbf{Step 2.2:} Use crossover and mutation to reproduce offspring $x^c$.
            \State \textbf{Step 2.3:} Update the subproblems.
        \EndFor
    \EndFor
    \State \textbf{Step 3:} Update $WV$ based on the solutions' fitness variances.
\EndWhile
\end{algorithmic}
\end{algorithm}

\subsection{PMAO-MOEA/D-ADF Framework}
% sayem
% \addc{It will be better if Sayem writes about the implementation details}

\subsubsection{Workflow}

\begin{figure}[!ht]
    \centering
    \includegraphics[width= 0.85\linewidth]{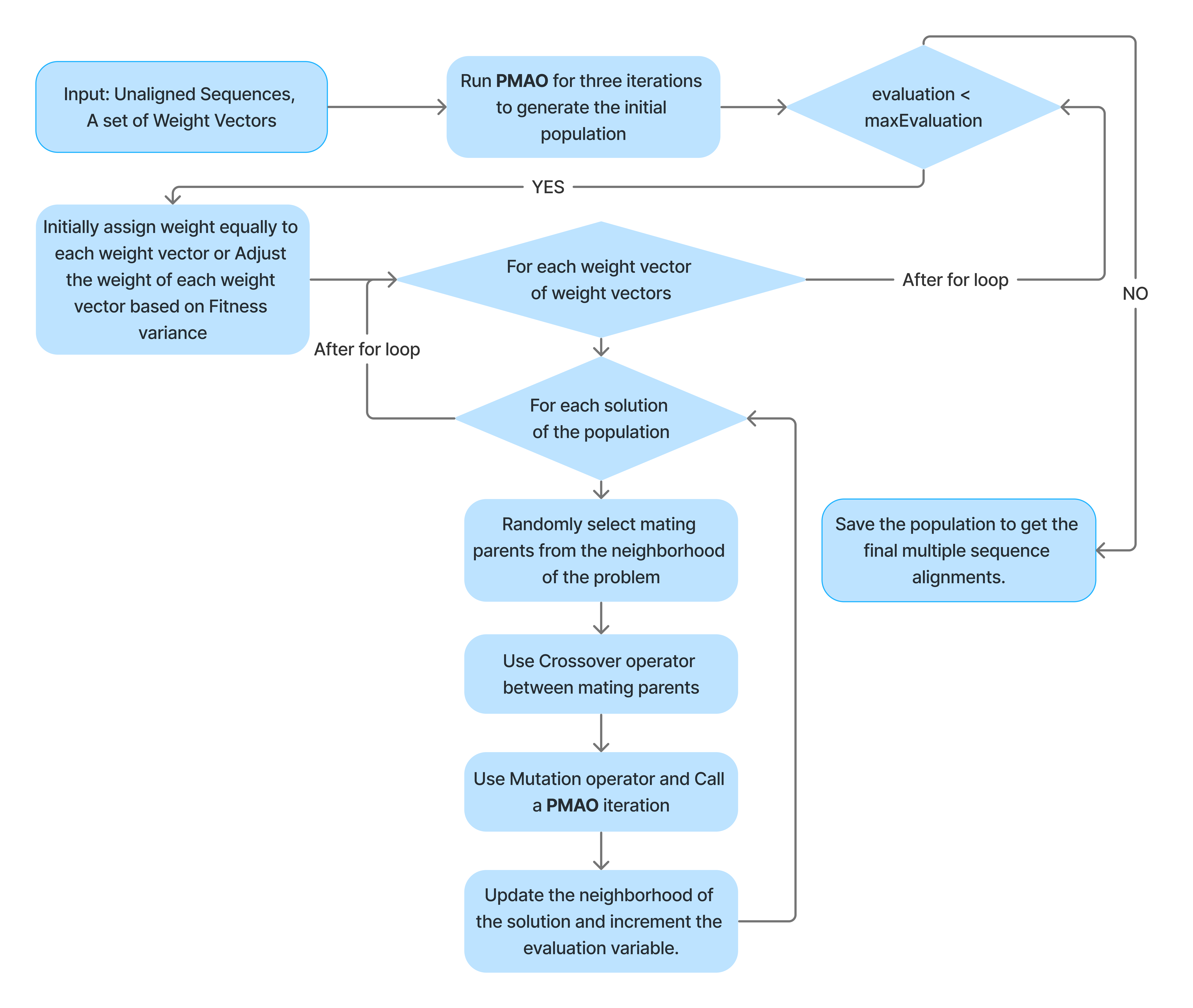}
    \caption{A high-level workflow of our PMAO-MOEA/D-ADF framework.}
    \label{fig:pmao-moead}
\end{figure}

\begin{enumerate}
    \item We apply the PMAO framework to construct the initial population, and this initial population contains 30 different solutions based on our choice of 30 weight vectors. 
    \item Initially, we assign an equal weight to each weight vector.
    \item while stopping criteria not met do
    \begin{enumerate}
        \item For each weight vector
    \begin{enumerate}
        \item For each solution of the population, 
        \begin{enumerate}
            \item We randomly select mating parents from the neighborhood of the solution.
            \item We apply the crossover operator (Single-Point Crossover) between mating parents.
            \item We use the mutation operator (Shift-closed gaps) to escape local optima by introducing small changes to solutions.
            \item We run a PMAO iteration to get a new (alignment, tree) pair from the solution. 
            \item We update the neighborhood. If the solution in the preceding step is superior to any solution in the neighborhood, this solution replaces the neighbor.
        \end{enumerate}
    \end{enumerate}
    \item We then update the weight of each weight vector based on the solutions' fitness variances.
    \end{enumerate}
    \item Finally, we save the population to collect the final set of non-dominated solutions that form the Pareto front, representing a trade-off between the defined objectives.
\end{enumerate}

\subsubsection{Input}
\begin{enumerate}
    \item \textbf{Unaligned Sequences:} A set of biological sequences (e.g., DNA, RNA, or protein sequences) that need to be aligned. These sequences are typically represented in raw, unaligned format.
    \item \textbf{Weight Vectors:} A set of \textbf{5D} weight vectors representing the relative importance of the objectives or criteria used for optimization. Each weight vector defines the trade-off between the objectives, guiding the multi-objective optimization process. These weight vectors help in selecting different solutions along the Pareto front, representing different trade-offs between objectives.
\end{enumerate}

\subsubsection{Output}
\textbf{A set of (MSA, Tree) Pairs:} The framework's main output is a set of pairs, each of which consists of a Multiple Sequence Alignment (MSA) and its related phylogenetic tree. These pairings illustrate the solutions produced by combining PMAO and MOEA/D-ADF.

\subsubsection{Weight Vectors}
The multi-objective optimization method is guided by a set of weight vectors. We use the same set of weight vectors as were used in the PMAO paper for our analysis. As described in the PMAO study, these weight vectors were carefully chosen to achieve a balance between computational efficiency and the exploration of different objectives. We aim to maintain continuity and facilitate direct comparisons with the results presented in the PMAO framework by using this consistent set of weight vectors, while also demonstrating the benefits of integrating with MOEA/D-ADF in the context of multiple sequence alignment.

\subsubsection{Initial Population}
We use the \textbf{PMAO} framework to initialize our multi-objective optimization framework, in which unaligned biological sequences and a predefined set of weight vectors are used as input parameters for the PMAO algorithm. We run the PMAO algorithm for \textbf{three} iterations, during which multiple sequence alignments (MSAs) and phylogenetic trees are co-estimated. We then extract the corresponding MSAs from PMAO's output. These high-quality MSAs are then identified as the initial population for our PMAO-MOEA/D-ADF framework. This method allows us to seed our optimization process with high-performing PMAO solutions, increasing the likelihood of convergence to superior solutions in the following optimization iterations.

\subsubsection{Initial Neighborhood}
Initializing the neighborhood involves choosing a collection of solutions in the objective space that are close to a given solution. The goal is to generate a neighborhood of solutions that possess similar qualities or compete with the initial solution. The following are the steps for finding neighbors:

\begin{enumerate}
    \item \textbf{Define a Reference Solution:} Select a reference solution from the population or dataset, usually the solution for which we want to find neighbors.

    \item \textbf{Calculate Distances:} For each of the other solutions, calculate the Euclidean distance between their objective values and the reference solution's objective values. 

    \item \textbf{Sort Solutions:} After calculating the Euclidean distances between the reference solution and all other solutions, sort the solutions in ascending order based on their distances. This sorting will identify the closest solutions in the objective space.

    \item \textbf{Select Neighbors:} Choose the top-k (where k is a predefined parameter) solutions with the smallest Euclidean distances as the nearest neighboring subproblems.
\end{enumerate}

\subsubsection{Crossover Operator}
We use the Single-Point Crossover. Single-Point Crossover randomly selects a position from parent A by splitting it into two blocks and parent B is tailored so that the right piece can be joined to the left piece of the first parent (PA1) and vice versa. Selected blocks are crossed between these two parents.

\subsubsection{Mutation Operator}
We use Shift-closed gaps as a mutation operator. Shift-Closed Gaps mutation operator randomly selects closed gaps (gaps without characters) in a sequence and shifts them to different positions, introducing diversity and altering the gap structure within the sequence.

\section{Experimental Results}\label{sec:results}

\subsection{Dataset}
We use the BAliBASE 3.0 benchmark dataset \cite{thompson2005balibase} to conduct all our experiments. It contains manually constructed, high-quality MSAs based on three-dimensional structural superpositions. It has 218 datasets categorized into six groups according to their families and similarities. These are:
\begin{enumerate}
    \item RV11: very divergent sequences, residue identity below 20\%
    \item RV12: medium to divergent sequences, with 20\%–40\%
residue identity
\item RV20: families with one or more highly divergent sequences
\item RV30: divergent subfamilies
\item RV40: sequences with large
terminal N/C extensions
\item RV50: sequences with large internal insertions
\end{enumerate}

\subsection{Performance Measure}
In this experiment, we have used FN rates as a performance measure. To evaluate the quality of each tree, we have used a widely accepted measure called the False Negative (FN) rate. The FN rate tells us the percentage of edges that are present in the true tree but are missing in the estimated tree. Therefore, a smaller value of the FN rate is desirable. It is worth noting that the other two common tree error measures, the False Positive rate and Robinson-Foulds rate, give the same measure as the FN rate when the true and estimated trees are binary trees \cite{warnow2017computational}.

\subsection{Results}
We will use the term PMAO++ to refer to our PMAO-MOEA/D-ADF method from now on, for simplicity. We have conducted a comparison between our method, PMAO++, and the state-of-the-art method, PMAO. To compare the performance of PMAO and PMAO++, we randomly selected 17 datasets from six categories of the BAliBASE 3.0 benchmark dataset. To test our method, we provided both PMAO and PMAO++ with 30 weight vectors and unaligned sequences. After obtaining the final aligned sequences and trees from both methods, we calculated the false negative rates (FN rates) to evaluate their performance.

We have compared the best and average FN rates on each dataset using both methods. On average FN rates, PMAO++ produced better FN rates in most of the datasets. Moreover, \textbf{PMAO++ outperformed the state-of-the-art method PMAO in 12 out of 17 datasets on average FN rates} (Table \ref{tab:pmao_vs_pmao++}). PMAO was only able to outperform PMAOO++ in four datasets. However, the average FN rate of PMAO++ is not significantly different from PMAO's FN rate in those four datasets.

It is worth noting that the best false negative (FN) rates reveal an interesting insight. PMAO++ performs better than PMAO in terms of best FN rate on most datasets (Table \ref{tab:pmao_vs_pmao++}), while PMAO outperforms PMAO++ only on one dataset. Therefore, this highlights the superiority of our method, PMAO++.

\begin{table}[!h]
\centering

\caption{Best and Average FN rates obtained by PMAO and PMAO++. Lower (better) FN rates have been marked with Glitchy Shader Blue and Light Periwinkle colors. Across the last two columns, the higher (better) counts have been marked with a greenish shade.}

\renewcommand{\arraystretch}{1.3}
\begin{tabular}{|c|cc|cc|cc|}
% \hline
%                                    & \multicolumn{2}{c|}{\textbf{Avg}}                                                  & \multicolumn{2}{c|}{\textbf{Best}}                                                & \multicolumn{2}{c|}{\textbf{Count}}                                          \\ 
%                                    \cline{2-7} 
% \multirow{-2}{*}{\textbf{Dataset}} & \multicolumn{1}{c|}{\textbf{PMAO}}                 & \textbf{PMAO++}               & \multicolumn{1}{c|}{\textbf{PMAO}}                & \textbf{PMAO++}               & \multicolumn{1}{c|}{\textbf{PMAO}}              & \textbf{PMAO++}            \\ \hline
\hline
\multirow{2}{*}{\textbf{Dataset}} & \multicolumn{2}{c|}{\textbf{Avg}} & \multicolumn{2}{c|}{\textbf{Best}} & \multicolumn{2}{c|}{\textbf{Count}} \\
% \cline{2-7} 
 & \textbf{PMAO} & \textbf{PMAO++} & \textbf{PMAO} & \textbf{PMAO++} & \textbf{PMAO} & \textbf{PMAO++} \\
\hline
\textbf{BB11007}                   & \multicolumn{1}{c|}{0.65}                          & \cellcolor[HTML]{96FFFB}0.58  & \multicolumn{1}{c|}{0.33}                         & 0.33                          & \multicolumn{1}{c|}{8}                          & \cellcolor[HTML]{9AFF99}16 \\ \hline
\textbf{BB11038}                   & \multicolumn{1}{c|}{0.60}                          & \cellcolor[HTML]{96FFFB}0.55  & \multicolumn{1}{c|}{0.4}                          & \cellcolor[HTML]{CBCEFB}0.0   & \multicolumn{1}{c|}{10}                         & \cellcolor[HTML]{9AFF99}12 \\ \hline
\textbf{BB11019}                   & \multicolumn{1}{c|}{0.31}                          & \cellcolor[HTML]{96FFFB}0.28  & \multicolumn{1}{c|}{0.14}                         & 0.14                          & \multicolumn{1}{c|}{8}                          & \cellcolor[HTML]{9AFF99}11 \\ \hline
\textbf{BB12012}                   & \multicolumn{1}{c|}{0}                             & 0                             & \multicolumn{1}{c|}{0}                            & 0                             & \multicolumn{1}{c|}{0}                          & 0                          \\ \hline
\textbf{BB12001}                   & \multicolumn{1}{c|}{\cellcolor[HTML]{96FFFB}0.30}  & 0.31                          & \multicolumn{1}{c|}{0.125}                        & 0.125                         & \multicolumn{1}{c|}{\cellcolor[HTML]{9AFF99}13} & 10                         \\ \hline
\textbf{BB12037}                   & \multicolumn{1}{c|}{0.55}                          & \cellcolor[HTML]{96FFFB}0.50  & \multicolumn{1}{c|}{0.3}                          & \cellcolor[HTML]{CBCEFB}0.2   & \multicolumn{1}{c|}{10}                         & \cellcolor[HTML]{9AFF99}16 \\ \hline
\textbf{BB20002}                   & \multicolumn{1}{c|}{0.68}                          & \cellcolor[HTML]{96FFFB}0.63  & \multicolumn{1}{c|}{0.58}                         & \cellcolor[HTML]{CBCEFB}0.47  & \multicolumn{1}{c|}{8}                          & \cellcolor[HTML]{9AFF99}18 \\ \hline
\textbf{BB20012}                   & \multicolumn{1}{c|}{0.30}                          & \cellcolor[HTML]{96FFFB}0.277 & \multicolumn{1}{c|}{0.21}                         & 0.21                          & \multicolumn{1}{c|}{9}                          & \cellcolor[HTML]{9AFF99}15 \\ \hline
\textbf{BB20030}                   & \multicolumn{1}{c|}{0.696}                         & \cellcolor[HTML]{96FFFB}0.656 & \multicolumn{1}{c|}{0.636}                        & \cellcolor[HTML]{CBCEFB}0.568 & \multicolumn{1}{c|}{6}                          & \cellcolor[HTML]{9AFF99}21 \\ \hline
\textbf{BB20037}                   & \multicolumn{1}{c|}{0.23}                          & \cellcolor[HTML]{96FFFB}0.196 & \multicolumn{1}{c|}{0.13}                         & 0.13                          & \multicolumn{1}{c|}{7}                          & \cellcolor[HTML]{9AFF99}21 \\ \hline
\textbf{BB30011}                   & \multicolumn{1}{c|}{0.35}                          & \cellcolor[HTML]{96FFFB}0.29  & \multicolumn{1}{c|}{0.30}                         & \cellcolor[HTML]{CBCEFB}0.225 & \multicolumn{1}{c|}{0}                          & \cellcolor[HTML]{9AFF99}30 \\ \hline
\textbf{BB30026}                   & \multicolumn{1}{c|}{\cellcolor[HTML]{96FFFB}0.20}  & 0.217                         & \multicolumn{1}{c|}{\cellcolor[HTML]{CBCEFB}0.15} & 0.19                          & \multicolumn{1}{c|}{\cellcolor[HTML]{9AFF99}18} & 5                          \\ \hline
\textbf{BB40009}                   & \multicolumn{1}{c|}{\cellcolor[HTML]{96FFFB}0.225} & 0.23                          & \multicolumn{1}{c|}{0.19}                         & 0.19                          & \multicolumn{1}{c|}{10}                         & 10                         \\ \hline
\textbf{BB40006}                   & \multicolumn{1}{c|}{\cellcolor[HTML]{96FFFB}0.067} & 0.087                         & \multicolumn{1}{c|}{0}                            & 0                             & \multicolumn{1}{c|}{\cellcolor[HTML]{9AFF99}13} & 6                          \\ \hline
\textbf{BB50002}                   & \multicolumn{1}{c|}{0.69}                          & \cellcolor[HTML]{96FFFB}0.676 & \multicolumn{1}{c|}{0.5}                          & 0.5                           & \multicolumn{1}{c|}{12}                         & \cellcolor[HTML]{9AFF99}16 \\ \hline
\textbf{BB50009}                   & \multicolumn{1}{c|}{0.23}                          & \cellcolor[HTML]{96FFFB}0.22  & \multicolumn{1}{c|}{0.08}                         & 0.08                          & \multicolumn{1}{c|}{11}                         & \cellcolor[HTML]{9AFF99}12 \\ \hline
\textbf{BB50014}                   & \multicolumn{1}{c|}{0.44}                          & \cellcolor[HTML]{96FFFB}0.36  & \multicolumn{1}{c|}{0.41}                         & \cellcolor[HTML]{CBCEFB}0.26  & \multicolumn{1}{c|}{1}                          & \cellcolor[HTML]{9AFF99}25 \\ \hline
\end{tabular}
\renewcommand{\arraystretch}{1}
\label{tab:pmao_vs_pmao++}
\end{table}

The final two columns in Table \ref{tab:pmao_vs_pmao++} indicate the number of trees with better FN rates for each method compared to the other method. PMAO++ generally outperforms PMAO on most datasets, resulting in higher counts for PMAO++. Thus, PMAO++ obtains the better tree overall on most datasets.

\section{Discussion}\label{sec:discussion}

% We performed a paired t-test to check the significance of our results. Our alternative hypothesis was that the mean of the obtained FN rates using PMAO++ is greater than the mean obtained using PMAO. The p-value we obtained was 0.64. This suggests that PMAO++ significantly improves the average FN rate and, hence, the overall quality of the population. It also indicates that PMAO++ is more effective in finding the best individual in most cases.

% Based on the comparison provided in Table \ref{tab:pmao_vs_pmao++}, it can be observed that PMAO++ did not perform better in terms of FN rates for the RV40 group. Although the difference between PMAO and PMAO++ is not significant, it is possible that PMAO++ may not be able to achieve better FN rates for sequences with large terminal N/C extensions. To confirm this hypothesis, further testing is required using additional datasets of the RV40 group.

% PMAO++ has demonstrated its ability to obtain better trees by achieving an FN rate of zero in some datasets. These results collectively highlight the robustness and effectiveness of PMAO++ in mitigating false negatives across diverse datasets. The consistent outperformance of PMAO++ in comparison to PMAO further underscores its potential as a superior solution. Additionally, the adaptability of PMAO++ to various dataset characteristics reinforces its applicability and usefulness in real-world scenarios.

% \section{Discussion}

This study presents MOEA/D-ADF and its integration with PMAO to form PMAO\texttt{++}, an objective-aware framework that produces ensembles of alignment--tree pairs for downstream phylogenetic analysis. Below we synthesize the methodological insights, biological implications, and practical limitations revealed by our experiments.

% \textbf{Biological relevance and utility.}  
PMAO\texttt{++} explicitly acknowledges alignment-induced uncertainty by returning many plausible alignments together with their corresponding trees. This ensemble output is directly useful to biologists: it can be fed to consensus and summary-tree pipelines, used to quantify topology uncertainty across alternative alignments, and applied in downstream tasks (ancestral state reconstruction, comparative genomics, molecular dating, or epidemiological reconstruction) in a way that reflects alignment variability. Because the framework is \emph{objective-aware}, practitioners may up-weight biologically important criteria (for example, secondary-structure conservation for rRNA, reduced gap penalties for closely related orthologs, or topology-focused objectives for outbreak tracing); the optimization will naturally incorporate such preferences while maintaining multi-objective trade-offs.

% \textbf{Algorithmic advantages.}  
Seeding an adaptive evolutionary search with PMAO solutions leverages high-quality starting points and allows MOEA/D-ADF to focus computational effort where uncertainty or diversity is greatest. The fitness-variance driven weight adaptation concentrates search in decomposition regions with high variance, improving exploration without sacrificing exploitation of promising basins. Practically, this improves Pareto-front coverage and increases the chance of discovering superior alignment--tree pairs, which we observed as consistently better FN rates on most benchmarks and several zero-FN best-case trees.

% \textbf{Dataset-specific limitations and remedies.}  
Certain sequence collections remain challenging: e.g., large terminal N/C extensions and atypical indel patterns (as in the RV40 group) can bias alignment scoring and hinder topology recovery. To address these issues we recommend (1) enriching the objective set with indel-aware or structural objectives (gap models informed by biology, secondary-structure agreement), (2) applying targeted preprocessing (trimming low-complexity termini when justified), and (3) incorporating model-based phylogenetic criteria (likelihood or model-fit measures) to better align alignment objectives with phylogenetic goals.

% \textbf{Practical considerations and future enhancements.}  
Scalability to large sequence collections is an important engineering challenge: parallelized fitness evaluations, more efficient neighborhood strategies, and incremental alignment techniques will be essential. Methodologically, hybridizing MOEA/D-ADF with local refinement heuristics, MCMC-based topology updates, or learned proposal mechanisms may further boost accuracy. Finally, user-facing tools—visualization of alignment--tree ensembles and automated summary-tree integration—will increase accessibility for domain scientists.

Collectively, these points indicate that PMAO\texttt{++} provides a flexible, biologically informed multi-objective approach that improves tree estimation in many realistic settings while offering clear paths for targeted improvements on difficult datasets.

\section{Conclusion}\label{sec:conclusion}

We introduced \textbf{MOEA/D-ADF}, a fitness-variance driven adaptive decomposition variant, and combined it with PMAO to create \textbf{PMAO\texttt{++}}. By seeding MOEA/D-ADF with PMAO-generated solutions and evolving with complementary weight vectors, PMAO\texttt{++} produces diverse alignment--tree ensembles that (1) improve alignment-derived phylogenies on a majority of tested benchmarks, (2) enable user-guided, objective-aware searches aligned with biological priorities, and (3) supply direct inputs for summary and consensus methods that better reflect alignment uncertainty.

Limitations include sensitivity to atypical sequence features (e.g., terminal extensions) and the computational cost of large-scale ensembles. Future work will focus on expanding objective sets (indel- and structure-aware criteria), integrating likelihood-based phylogenetic measures, improving scalability, and building visualization and pipeline tools to translate alignment--tree ensembles into actionable biological inferences.

%%==================================%%
%% Declarations
%%==================================%%

\backmatter

% \bmhead{Supplementary information}

% The online version contains supplementary material available at \url{https://doi.org/10.1038/s123456}.

% \bmhead{Acknowledgments}

% We would like to thank the anonymous reviewers for their valuable comments and suggestions.

\section*{Declarations}

\subsection*{Ethics approval and consent to participate}
Not applicable.

\subsection*{Consent for publication}
Not applicable.

\subsection*{Code, Environment and Availability}

\textbf{Environment:} You can find all instructions to install and run the experiments at \url{https://github.com/SaemHasan/MOEAD-Extension/blob/main/README.md}. The requirements to run the project are:
\begin{enumerate}
    \item Java SE Development Kit 8
    \item Apache Maven
    \item Python 3.10
    \item Python 2.7
\end{enumerate}

We conducted the experiment on Ubuntu 20.04 LTS OS, running on an Intel 13th Gen Core i5 processor and a GeForce RTX 3060Ti 8GB GPU.

\textbf{Availability:} You can find the codebase and datasets at \url{https://github.com/SaemHasan/MOEAD-Extension}.

\subsection*{Competing interests}
The authors declare that they have no competing interests.

\subsection*{Funding}
The authors received no financial support for this research.

% \bmhead{Authors' contributions}
% S.H. conceived the study, implemented the algorithm, conducted the experiments, and wrote the manuscript. M.A.N. provided guidance on experimental design and reviewed the manuscript. M.S.R. supervised the project, provided theoretical insights, and reviewed the manuscript. All authors read and approved the final manuscript.

%%==================================%%
%% Bibliography
%%==================================%%

% \bibliographystyle{sn-mathphys-num}
\bibliography{cas-refs}

\end{document}